\documentclass{article}
\usepackage[preprint]{corl_2026} 

\usepackage{xcolor}
\usepackage{booktabs}   
\usepackage{graphicx}   
\usepackage{tabularx}

\newif\ifteacheredits
\ifdefined\showteacheredits
  \teachereditstrue
\else
  \teachereditsfalse
\fi
\DeclareRobustCommand{\teacheredit}[1]{\ifteacheredits{\color{red}#1}\else#1\fi}
\newcommand{\teachereditcolor}{\ifteacheredits\color{red}\fi}

\definecolor{topogreen}{RGB}{0,128,0}
\newcommand{\topomethod}{\textbf{\textcolor{topogreen}{TopoRetarget}}}
\usepackage{amsmath}

\usepackage{amsfonts}

\title{\topomethod{}: \teacheredit{Interaction-Preserving Retargeting for Dexterous Manipulation}}

\author{
  \textbf{Jielin Wu$^{*}$, Shenzhe Yao$^{*}$, Guanqi He$^{*\ddagger}$, Xiaohan Liu$^{*\ddagger}$, Zhaoqing Zeng, Xiangrui Jiang,}\\
  \textbf{Han Yang, Wentao Zhang, Hang Zhao$^{\dagger}$}\\
  \normalfont IIIS, Tsinghua University\\
  \normalfont $^{*}$Equal contribution \quad $^{\ddagger}$Project lead \quad $^{\dagger}$Corresponding author\\
  \normalfont \texttt{Project Page:}~\href{https://toporetarget2026.github.io/TopoRetarget/}{\texttt{https://toporetarget2026.github.io/TopoRetarget/}}
}

\begin{document}
\vspace*{-15mm}

\maketitle

\vspace{-10mm}
\begin{figure}[htbp]
    \centering
     \includegraphics[width=0.9\textwidth]{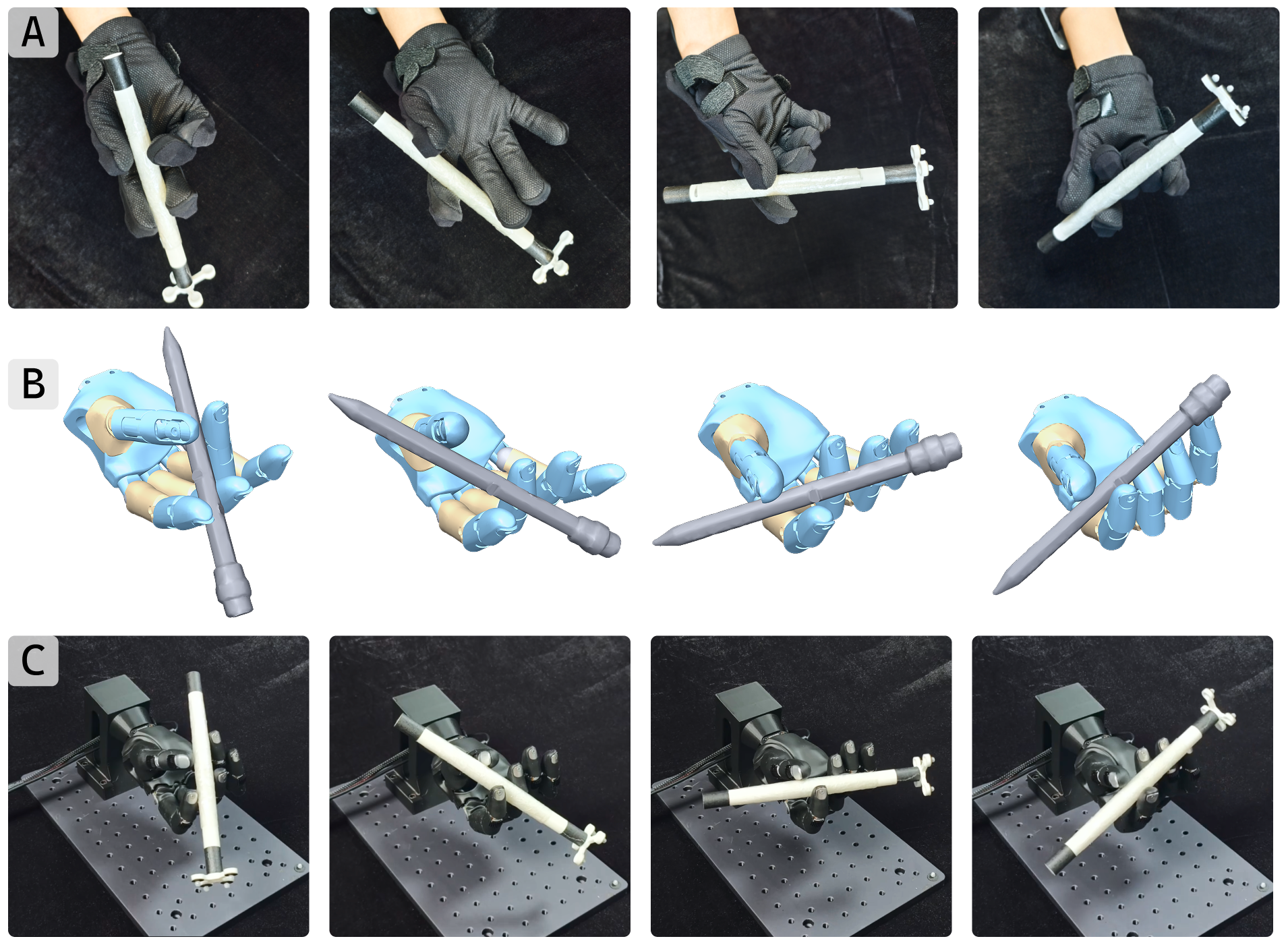}
 \caption{\topomethod{} converts human hand-object trajectories collected with a motion-capture glove (A) into robot references for contact-rich dexterous manipulation.
These retargeted references are used to train RL policies that execute dexterous skills in simulation (B) and transfer zero-shot to Wuji Hand (C).}
    \label{fig:teaser}
\end{figure}


\begin{abstract}
Human hand-object demonstrations provide dense reference motions for training dexterous manipulation reinforcement learning (RL) policies through reference tracking.
However, to use such demonstrations for \teacheredit{RL} policy learning, retargeting must preserve
hand pose and \teacheredit{task-relevant hand-object contact structure.}
Otherwise, contact and feasibility artifacts can \teacheredit{degrade downstream RL policy performance.}
We introduce \topomethod{}, an \teacheredit{interaction-preserving} retargeting framework that uses \teacheredit{a single set of parameters} across diverse retargeting conditions
while \teacheredit{maintaining task-relevant hand-object interaction} and adapting human demonstrations to dexterous robot hands.
The method constructs a sparse interaction graph over hand and object keypoints and optimizes
distance-weighted Laplacian deformation with directional consistency, kinematic constraints,
and penetration handling.
Evaluations show that the generated references improve both interaction fidelity and policy learning:
\topomethod{} achieves the best contact precision and alignment over all baselines
on the ContactPose Dataset, improves Pen-Spin training success by 40.6 percentage points over the
existing baseline methods, and enables zero-shot transfer to Wuji Hand hardware on cube reorientation and
pen spinning.
Project website: \href{https://toporetarget2026.github.io/TopoRetarget/}{\textcolor{blue}{TopoRetarget}}.




\end{abstract}

\keywords{motion retargeting, reinforcement learning, dexterous manipulation, trajectory optimization} 


\section{Introduction}

Learning long-horizon, contact-rich dexterous manipulation with reinforcement learning often relies on carefully designed training pipelines~\cite{andrychowicz2020learning,handa2023dextreme,wang2024lessons}. Human hand-object demonstrations provide dense reference signals that guide reinforcement learning through tracking objectives~\cite{liu2025dextrack,xu2025dexplore,mandi2025dexmachina,li2025maniptrans}. Using these demonstrations for robot policy learning requires retargeting human hand-object motion across a substantial embodiment gap, but matching hand pose alone is insufficient since task execution depends on local hand-object interaction. Therefore, preserving hand-object interaction during retargeting is a key bottleneck for learning dexterous manipulation skills from human demonstrations.



Existing dexterous hand retargeting methods have made human hand motion useful for teleoperation and demonstration collection~\cite{handa2020dexpilot,qin2023anyteleop,sivakumar2022robotic,qin2022one}. Most of these pipelines define hand-centric objectives over hand pose~\cite{xin2026analyzing}, fingertip correspondences~\cite{handa2020dexpilot}, and pinch relations~\cite{yin2025geometric}. Such objectives align the robot hand with the human hand, but do not directly specify where and how the robot should contact the object. Recent methods address this limitation by incorporating object motion~\cite{chen2024object,mandi2025dexmachina,li2025maniptrans}, contact consistency~\cite{lin2025dexflow,huang2025fungrasp}, or physical feasibility~\cite{pan2025spider} into retargeting or policy learning. However, the resulting motions often still require downstream refinement, policy correction, or task-specific tuning before they can serve as reliable tracking references.

This motivates \topomethod{}, a retargeting framework that focuses on the structure of the local hand-object interaction. It preserves object-relative positional and directional relationships between hand links and object regions, while also maintaining intra-hand relationships among hand links. Our approach reduces interaction artifacts and supports real-time retargeting.







Our contributions are threefold:
\begingroup
\setlength{\leftmargini}{1.5em}
\begin{enumerate}
    \item An \teacheredit{interaction-preserving} retargeting framework that \teacheredit{maintains local hand-object interaction} while enforcing dexterous hand kinematic and penetration constraints.
    \item A lightweight reference-based RL tracking pipeline that enables zero-shot sim-to-real transfer for contact-rich dexterous skills, including pen spinning and cube reorientation.
    \item A dexterous hand-object interaction dataset of retargeted trajectories, task references, and trained policies for reproducible reference-based dexterous manipulation.
\end{enumerate}
\endgroup
\section{Related Work}
     
\subsection{Dexterous Hand Motion Retargeting}

Dexterous hand motion retargeting maps human hand observations to robot hand motion for teleoperation and demonstration collection. DexPilot formulates vision-based dexterous hand-arm teleoperation around task-space hand objectives~\cite{handa2020dexpilot}, and AnyTeleop extends vision-based hand-arm teleoperation across different robotic embodiments~\cite{qin2023anyteleop}. Later studies analyze how hand-pose, fingertip, keyvector, joint-limit, smoothness, and collision objectives affect retargeting quality~\cite{xin2026analyzing}, while GeoRT proposes an ultrafast geometric retargeting model for hand motion transfer~\cite{yin2025geometric}. These methods enable practical teleoperation and demonstration collection, but their objectives remain largely hand-centric. For contact-rich policy learning, references must preserve object-relative local interaction, including contacts on fingertips, intermediate phalanges, finger sides, and the palm. DexFlow incorporates hand-object interaction into hand pose retargeting~\cite{lin2025dexflow}, FunGrasp focuses on functional grasping for diverse dexterous hands~\cite{huang2025fungrasp}, and GenHand studies generalized human grasp kinematic retargeting~\cite{qi2026genhand}. These methods improve grasp or contact transfer, but frequent contact-mode transitions still require references that preserve evolving hand-object relations. \topomethod{} treats retargeting as interaction-preserving reference generation, \teacheredit{maintaining local hand-object interaction} while enforcing robot kinematics and penetration constraints.

\subsection{Reinforcement Learning for Dexterous Manipulation}

Reinforcement learning for dexterous manipulation trains policies for high-dimensional contact-rich skills through task-specific objectives, domain randomization, curriculum learning, and sim-to-real adaptation. OpenAI demonstrates cube reorientation with large-scale simulation and domain randomization~\cite{andrychowicz2020learning}; DeXtreme transfers agile in-hand manipulation from simulation to real hardware~\cite{handa2023dextreme}; RMA-based in-hand rotation adapts to object and dynamics variations through learned adaptation~\cite{qi2023hand}; and recent pen-spinning systems still require carefully designed training pipelines or real-world trial-and-error search for dynamic multi-finger contact~\cite{wang2024lessons,yao2025softspin}. To reduce this burden, reference-based learning turns long-horizon exploration into trajectory tracking. DeepMimic establishes this paradigm for physics-based character skills~\cite{peng2018deepmimic}, and OmniH2O applies human-to-humanoid motion tracking to whole-body control~\cite{he2024omnih2o}. Dexterous-hand systems have begun to adopt reference-guided policy learning: DexTrack trains neural tracking control from human references~\cite{liu2025dextrack}, Dexplore uses reference-scoped exploration~\cite{xu2025dexplore}, DexMachina studies functional retargeting for bimanual dexterous manipulation~\cite{mandi2025dexmachina}, and ManipTrans uses residual learning for dexterous bimanual transfer~\cite{li2025maniptrans}. Reference-based learning depends directly on reference quality: policies must be able to follow the provided references. \topomethod{} targets this bottleneck by producing references that preserve task-relevant local interaction before RL training.

\subsection{Interaction-Preserving Data Generation}

\teacheredit{Interaction-preserving data generation} turns human hand-object behavior into policy references that retain task-relevant spatial relationships. DexMV and Robotic Telekinesis recover motion from human videos~\cite{qin2022dexmv,sivakumar2022robotic}; From One Hand to Multiple Hands uses single-camera teleoperation for multi-hand imitation~\cite{qin2022one}; DexCap provides portable MoCap collection~\cite{wang2024dexcap}; DexMimicGen scales a small set of human demonstrations into large bimanual dexterous datasets~\cite{jiang2025dexmimicgen}; and the ContactPose Dataset provides hand-object contact annotations~\cite{brahmbhatt2020contactpose}. These sources collect hand-object motion, but embodiment gaps can still produce contact misalignment, penetration, or infeasible robot poses. ObjDex guides policy learning with object goals and human wrist motion~\cite{chen2024object}, SPIDER uses physics-based sampling and virtual contact guidance for feasible dexterous trajectories~\cite{pan2025spider}, and OmniRetarget preserves \teacheredit{robot-object} relationships through contact-aware mesh deformation~\cite{yang2025omniretarget}. \topomethod{} provides real-time retargeting with local contact fidelity using \teacheredit{a fixed set of parameters} across dexterous in-hand manipulation settings.

\section{\topomethod{}: \teacheredit{Interaction-Preserving} Hand-Object Retargeting}

\begin{figure}[t]
    \centering
    \includegraphics[width=\textwidth]{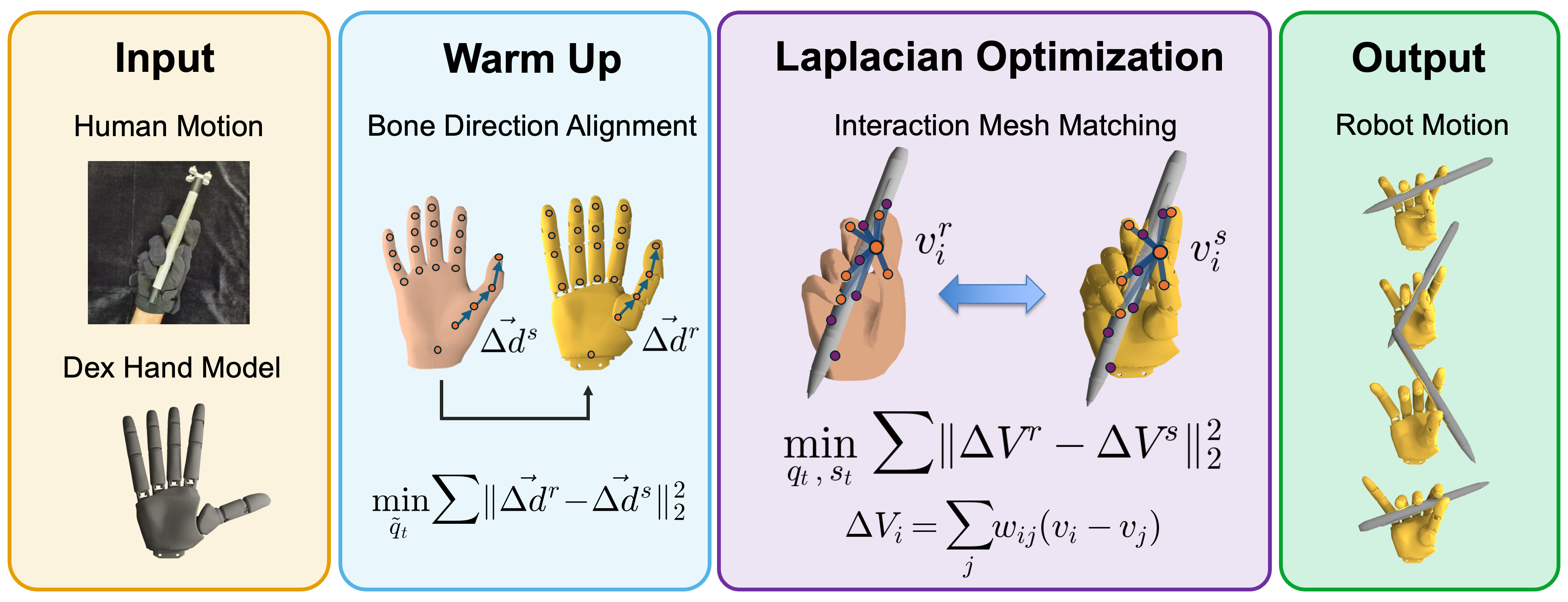}
    \caption{ \topomethod{} overview. Given a human demonstration, object mesh, and target hand model, the method aligns bone directions during initialization, constructs source and robot interaction meshes, and computes the robot configuration via topology-aware Laplacian optimization. The output robot motion reference preserves hand-object interaction.}
    \label{fig:system_overview}
\end{figure}

We propose \topomethod{}, an \teacheredit{interaction-preserving} retargeting framework for converting human hand-object demonstrations into robot references. The algorithm overview is shown in Fig.~\ref{fig:system_overview}. We first define the notation and optimization variables in Sec.~\ref{sec:method_notation}, then describe the relative bone-direction initialization in Sec.~\ref{sec:warm_start}, the shared hand-object interaction mesh in Sec.~\ref{sec:interaction_mesh}, and the topology-aware Laplacian optimization in Sec.~\ref{sec:topology_laplacian}.

\subsection{Notation Definition}
\label{sec:method_notation}

We formulate hand-object retargeting as a constrained optimization problem. The inputs are a human-hand trajectory $P^h_{1:N}$, an object pose trajectory $q^o_{1:N}$, and an object mesh $\mathcal{M}$.  Here, $P^h_t\in\mathbb{R}^{21\times 3}$ denotes the MediaPipe~\cite{lugaresi2019mediapipe} hand keypoints at time $t$, and $N$ denotes the trajectory length. The goal is to compute an optimal robot base-pose trajectory $q^{\mathrm{base}}_{1:N}$ and a robot joint trajectory $q^\theta_{1:N}\in\mathbb{R}^{N\times n_\theta}$. We write $q_t^r=(q_t^{\mathrm{base}},q_t^\theta)$ and use $q$ for $q_t^r$ in single-frame objectives.

\subsection{Relative Bone-Direction Initialization}
\label{sec:warm_start}

We first compute an initial guess for the robot-hand configuration by matching local finger articulation between the source and robot hands. For each non-terminal finger keypoint $k$, define $d_k$ as the unit vector from keypoint $k$ to its child keypoint along the same finger. Let $d_k^s$ and $d_k^r(q)$ denote the source and robot bone directions at frame $t$, expressed in their respective wrist-centered hand frames, and let $\mathcal{A}_B$ contain adjacent bone pairs along each finger. The bone-direction mismatch is defined as

\vspace{-3mm}
\begin{equation}
E_{\mathrm{bone}}(q)=\sum_{(k,l)\in\mathcal{A}_B}\left\|\left(d_k^r(q)-d_l^r(q)\right)-\left(d_k^s-d_l^s\right)\right\|_2^2 .
\end{equation}

The initial estimate for frame $t$ is
\begin{equation}
\tilde{q}_t^r:=\arg\min_q \lambda_{\mathrm{warm}}E_{\mathrm{bone}}(q)+\lambda_{\mathrm{smooth}}\left\|q-\tilde{q}_{t-1}^r\right\|_2^2 .
\end{equation}
The first term encourages the robot hand to reproduce the source hand's relative bone directions, while the second term promotes temporal consistency with the previous initial estimate. The scalar weights $\lambda_{\mathrm{warm}}$ and $\lambda_{\mathrm{smooth}}$ balance these two objectives. We use $\lambda_{\mathrm{warm}}$ for the initialization-stage bone-direction weight.

\subsection{Interaction Mesh Construction}
\label{sec:interaction_mesh}

At each frame $t$, we build source and robot hand-object vertex sets from the human keypoints $P_t^h$, the corresponding robot keypoints $P_t^r(q)$, and $N_o$ object surface samples $O_t$ drawn from the mesh $\mathcal{M}$ posed by $q_t^o$:
\begin{equation}
V_t^s=[P_t^h;O_t], \qquad V_t^r(q)=[P_t^r(q);O_t].
\end{equation}
Here, $N_v=21+N_o$ is the total number of graph vertices. We write $V_t^s=\{v_{i,t}^s\}_{i=1}^{N_v}$ and $V_t^r(q)=\{v_{i,t}^r(q)\}_{i=1}^{N_v}$, where the first 21 vertices are hand keypoints and the remaining vertices are object samples.
We run Delaunay tetrahedralization on the source vertices $V_t^s$ to obtain the interaction edge set $\mathcal{I}_t$, then reuse the same connectivity for the robot vertices:
\begin{equation}
G_t^s=(V_t^s,\mathcal{I}_t), \qquad G_t^r(q)=(V_t^r(q),\mathcal{I}_t).
\end{equation}
This shared graph lets the refinement compare the demonstrated and retargeted configurations under the same local hand-object neighborhood structure.
It also supports augmentation across object scales and hand embodiments: after scaling or replacing the object, $O_t$ follows the new surface, and the rebuilt graph lets the same Laplacian refinement preserve local interaction without new contact targets.

\subsection{Topology-Aware Laplacian Optimization}
\label{sec:topology_laplacian}

Starting from $\tilde{q}_t^r$, we refine the robot configuration by matching topology-aware Laplacian coordinates on the shared \teacheredit{interaction} graph. Let $\mathcal{N}_t(i)$ be the neighbors of vertex $i$ in $\mathcal{I}_t$. For each interaction edge $(i,j)\in\mathcal{I}_t$, we compute source-frame distance-aware weights
\begin{equation}
\tilde{w}_{ij,t}=\exp\left(-\kappa\left\|v_{i,t}^s-v_{j,t}^s\right\|_2\right), \qquad
w_{ij,t}=\frac{\tilde{w}_{ij,t}}{\sum_{j'\in\mathcal{N}_t(i)}\tilde{w}_{ij',t}} .
\end{equation}
Here, $\kappa$ is a spatial decay parameter. The weights are computed on the source configuration and reused for the robot configuration.
For a vertex set $V=\{v_i\}_{i=1}^{N_v}$, its weighted Laplacian coordinate at vertex $i$ is
\begin{equation}
\Delta_t(V)_i=\sum_{j\in\mathcal{N}_t(i)}w_{ij,t}(v_i-v_j).
\end{equation}
We define the interaction-mesh energy as
\begin{equation}
E_{\mathrm{IM}}(q)=\frac{1}{N_v}\sum_{i=1}^{N_v}\left\|\Delta_t(V_t^r(q))_i-\Delta_t(V_t^s)_i\right\|_2^2 .
\end{equation}
The final interaction-preserving optimization is
\begin{equation}
\begin{aligned}
(q_t^{r,*},s_t^*)=\arg\min_{q,s}\quad&
\lambda_{\mathrm{IM}}E_{\mathrm{IM}}(q)
+\lambda_{\mathrm{bone}}E_{\mathrm{bone}}(q)
+E_{\mathrm{reg}}(q;q_{t-1}^{r,*})
+\frac{w_s}{2}\sum_{i\in\mathcal{Q}_t} s_i^2 \\
\mathrm{s.t.}\quad&
\phi_i(q)\ge -b,\qquad
\phi_i(q)+s_i\ge -\tau,\qquad
0\le s_i\le b-\tau,\qquad i\in\mathcal{Q}_t .
\end{aligned}
\end{equation}
Here, the $\lambda$ terms are scalar weights, $E_{\mathrm{IM}}$ \teacheredit{maintains the demonstration's local hand-object interaction}, $E_{\mathrm{bone}}$ keeps the initialization bone-direction prior, and $E_{\mathrm{reg}}$ regularizes smoothness. $\lambda_{\mathrm{bone}}$ denotes the refinement-stage bone-direction prior weight. The implementation of $E_{\mathrm{reg}}$ and the fixed parameter values are given in Appendix~\ref{app:retarget_details}. We use one fixed parameter setting across our experiments rather than performing extensive per-case tuning; detailed parameter values are provided in Table~\ref{tab:supp_retarget_params}. The query set $\mathcal{Q}_t$ contains hand-object pairs checked at frame $t$; for pair $i$, $\phi_i(q)$ is the signed distance, $s_i$ is the slack variable, $\tau$ is the soft tolerance, $b$ is the hard bound, and $w_s$ is the slack penalty weight.







\section{Minimal RL Tracking Controller}
\label{sec:rl_tracking}

We use a minimal reinforcement-learning setup to track the retargeted trajectories. We formulate the hand-object tracking task as a finite-horizon Markov decision process (MDP) and optimize the policy with Proximal Policy Optimization (PPO)~\cite{schulman2017proximal}. Each reference clip is $\xi^{\mathrm{ref}}=\{q^{\theta,\mathrm{ref}}_k,\;q^{o,\mathrm{ref}}_k,\;p^{\mathrm{ref}}_{1:L,k}\}_{k=0}^{N-1}$, where $q^{\theta,\mathrm{ref}}_k$ denotes the joint component $q_k^{\theta,*}$ of the retargeted configuration $q_k^{r,*}$, $q^{o,\mathrm{ref}}_k$ denotes the object pose expressed in the robot base frame, and $p^{\mathrm{ref}}_{1:L,k}$ denotes the base-frame positions of the $L$ tracked hand links. Expressing object and link references in the base frame makes the controller focus on relative hand-object interaction rather than global motion.

\noindent\textbf{Action space.}
At each control step $t$, the policy tracks the active reference frame $k_t$ using a residual action, $q_t^{\theta,\mathrm{tar}}=q_{k_t}^{\theta,\mathrm{ref}}+a_t$.

\noindent\textbf{Observation.}
The observation is $o_t=[o_t^{\mathrm{prop}},o_t^{\mathrm{obj}},o_t^{\mathrm{ref}}]$, where $o_t^{\mathrm{prop}}$ contains the joint state and previous action, $o_t^{\mathrm{obj}}$ contains object-axis points, and $o_t^{\mathrm{ref}}$ contains lookahead joint, object, and hand-link references in the robot base frame.

\noindent\textbf{Reference initialization and sampling.}
We use reference-state initialization: at reset, a start frame $k_0$ is sampled uniformly from the reference trajectory, and each environment is initialized with the corresponding hand and object states.

\noindent\textbf{Reward.}
The reward is $r_t=w_{\mathrm{obj}}r_{\mathrm{obj}}+w_{\mathrm{link}}r_{\mathrm{link}}+w_{\mathrm{joint}}r_{\mathrm{joint}}+w_{\mathrm{smooth}}r_{\mathrm{smooth}}$, where the four terms track object pose, hand-link positions, finger joints, and smoothness, respectively; the weights are listed in Appendix~\ref{app:rl_details}.

\noindent\textbf{Domain randomization.}
During training, we randomize object mass and center of mass, contact parameters, actuator gains, joint damping, and hand-link inertial properties, and apply external object perturbations during rollout.

\section{Experiments}






Our experiments evaluate the proposed retargeting pipeline through three questions:
\begingroup
\setlength{\leftmargini}{1.5em}
\begin{enumerate}
    \item Does \topomethod{} preserve local hand-object interactions across diverse human motions?
    \item How do retargeted references from different methods affect downstream RL tracking policy training and performance?
    \item Does \topomethod{} support generalization and augmentation across different objects and robot hands without per-case retuning?
\end{enumerate}
\endgroup
Across the retargeting and downstream tracking experiments, we compare the proposed method with four baselines: OmniRetarget~\cite{yang2025omniretarget} \teacheredit{(see details at Appendix~\ref{app:omniretarget_baseline})}, Mink~\cite{Zakka_Mink_Python_inverse_2026}, DexPilot~\cite{handa2020dexpilot}, and GeoRT~\cite{yin2025geometric}. All applicable methods use the same human references and object geometry.

\subsection{Interaction-Preserving Retargeting Quality}
\label{sec:retarget_quality}
To answer Q1, we compare different retargeting algorithms in terms of their ability to preserve local
  contact information in hand-object interaction.
We treat source in-contact bone segments as the key interaction units and \teacheredit{evaluate preservation of object-relative positions and directions after retargeting}. We report two metrics on the ContactPose Dataset~\cite{brahmbhatt2020contactpose}: $E_{\mathrm{prec}}$, the mean object-relative position displacement of contact links, and $E_{\mathrm{align}}$, the mean object-relative orientation error of contact links. Lower values indicate better interaction preservation. Full definitions are given in Appendix~\ref{app:retarget_metrics}.

Table~\ref{tab:headline_t1_comparison} shows that \topomethod{} achieves the lowest contact precision and contact alignment errors, with 7.71\,mm and 15.67$^\circ$, respectively. The residual is attributable to the human-dexterous embodiment gap and source-side ContactPose interpenetration. \topomethod{} also keeps penetration small, with a maximum penetration of 1.07\,mm, and avoids the severe interpenetration observed in several baselines, indicating that the method better preserves local contact geometry. 
Moreover, the average solve time is below 5\,ms per frame, enabling real-time retargeting.


\begin{table}[htbp]
    \centering
    \caption{Quantitative retargeting evaluation on the ContactPose Dataset~\cite{brahmbhatt2020contactpose}}
    \label{tab:headline_t1_comparison}
    \resizebox{\linewidth}{!}{%
    \begin{tabular}{lccccc}
    \toprule
    \textbf{Metric} & \topomethod{} (Ours) & OmniRetarget & Mink & DexPilot & GeoRT \\
    \midrule
    Contact precision (mm)\,$\downarrow$                       & \textbf{7.71}  & 14.15 & 14.12 & 14.13  & 26.77 \\
    Contact alignment ($^{\circ}$)\,$\downarrow$               & \textbf{15.67} & 30.80 & 37.36 & 33.71  & 25.74 \\
    Max hand-object penetration (mm)\,$\downarrow$             & \textbf{1.07}  & 1.15  & 20.12 & 11.87  & 22.22 \\
    \% frames with penetration $>$ 2\,mm\,$\downarrow$         & \textbf{0.00}  & \textbf{0.00} & 84.00 & 88.00  & 96.00 \\
    Solve time (ms\,/\,frame)\,$\downarrow$                    & 4.70           & 40.96 & 4.37  & 1.74   & \textbf{1.17} \\
    \bottomrule
    \end{tabular}}
\end{table}

\begin{figure}[htbp]
    \centering
    \vspace{-3mm}
    \includegraphics[width=0.95\textwidth]{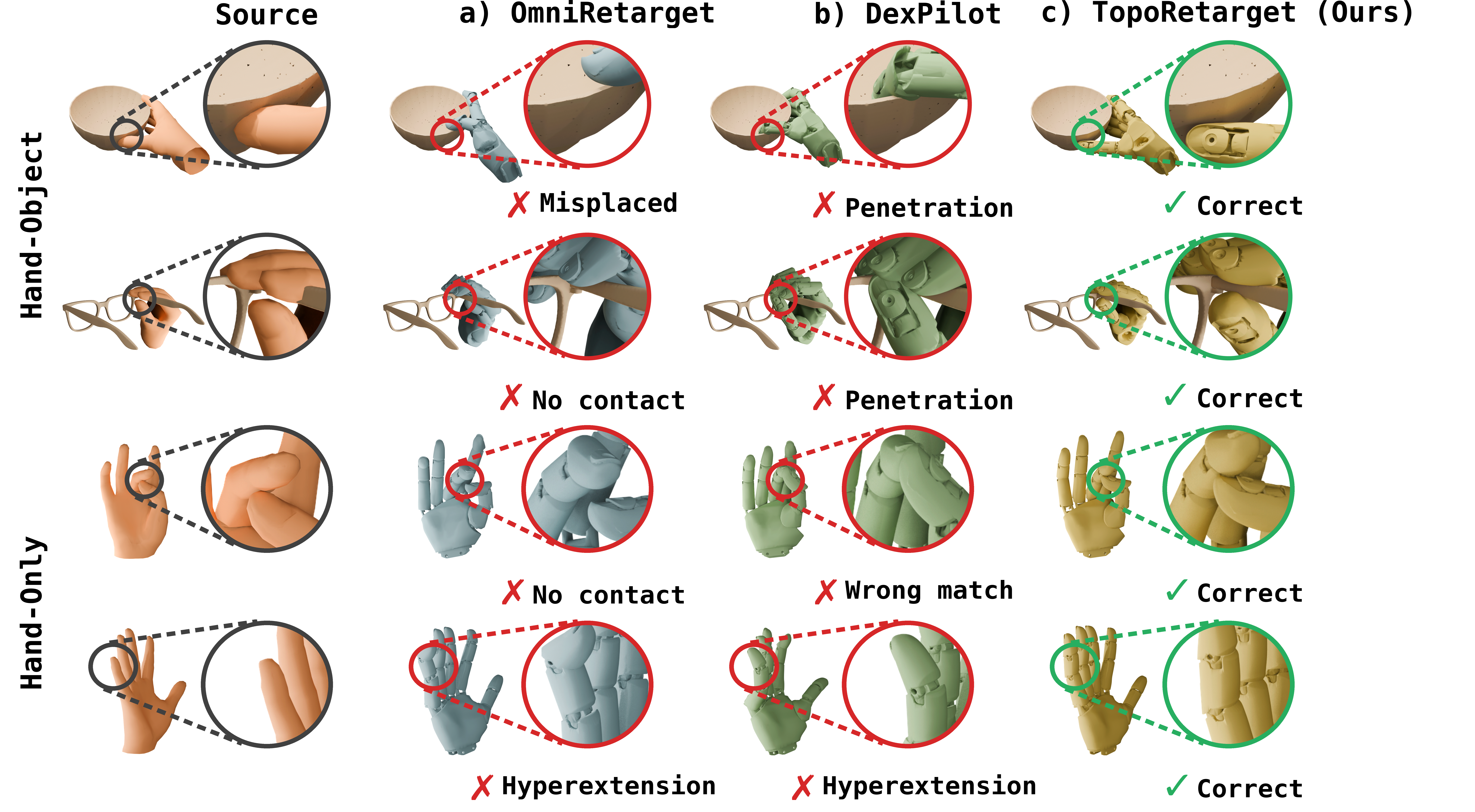}
\vspace{-2mm}
\caption{Retargeting artifacts of existing methods under hand-object and hand-only cases.}

    \label{fig:retarget_quality}
\end{figure}
\vspace{-2mm}




Fig.~\ref{fig:retarget_quality} summarizes common retargeting failure modes, including contact-region drift, invalid grasps or penetration, misplaced non-tip contacts, and unnatural joint configurations. In contrast, \topomethod{} preserves the hand-object interaction of the source while reducing artifacts. Detailed examples are shown in the figure.





\subsection{Downstream Tracking Policy Performance}
\label{sec:tracking_policy}

To answer Q2, we use downstream RL tracking policy training as a practical test of trajectory quality. For each retargeting method,
we generate references on the Ho-cap~\cite{wang2026ho} and our self-collected MoCap Pen-Spin Dataset (see details in Appendix~\ref{sec:experimental_setup}).
We then train RL tracking policies with the setup described in Sec.~\ref{sec:rl_tracking}, using identical policy architecture, reward terms, domain randomization, and termination criteria.
We evaluate the trained policies in simulation and report success rate and object tracking error. A rollout is counted as successful if it reaches the final reference frame without triggering the object-pose termination criteria in Appendix~\ref{app:rl_details}. 





\begin{table}[htbp]
    \centering
    \caption{Downstream RL tracking performance}
    \label{tab:retarget_tracking_combined}
    \footnotesize
    \setlength{\tabcolsep}{3pt}
    \renewcommand{\arraystretch}{1.05}
    \resizebox{\linewidth}{!}{%
    \begin{tabular}{lccccc}
        \toprule
        Metric & \topomethod{} (Ours) & OmniRetarget & Mink & DexPilot & GeoRT \\
        \midrule
        \multicolumn{6}{l}{\textit{Ho-cap~\cite{wang2026ho} Dataset (32 clips)}} \\
        \midrule
        Success rate\,$\uparrow$    & \textbf{84.4\% (27/32)} & 56.2\% (18/32) & 75.0\% (24/32) & 75.0\% (24/32) & 75.0\% (24/32) \\
        Obj. pos err (cm)\,$\downarrow$  & \textbf{0.87}           & 1.07           & 0.91           & 0.92           & 0.90 \\
        Obj. rot err (deg)\,$\downarrow$ & 5.76                    & 7.87           & 5.55           & \textbf{4.99}  & 6.07 \\
        \midrule
        \multicolumn{6}{l}{\textit{MoCap Pen-Spin Dataset (32 clips)}} \\
        \midrule
        Success rate\,$\uparrow$    & \textbf{87.5\% (28/32)} & 46.9\% (15/32) & 21.9\% (7/32)  & 40.6\% (13/32) & 31.2\% (10/32) \\
        Obj. pos err (cm)\,$\downarrow$  & \textbf{0.98}           & 1.45           & 1.61           & 1.29           & 1.19 \\
        Obj. rot err (deg)\,$\downarrow$ & \textbf{9.25}           & 14.26          & 15.25          & 17.66          & 18.62 \\
        \bottomrule
    \end{tabular}
    }
    \vspace{-3mm}
\end{table}

\vspace{-5mm}
\begin{figure}[htbp]
    \centering
    \includegraphics[width=0.95\textwidth]{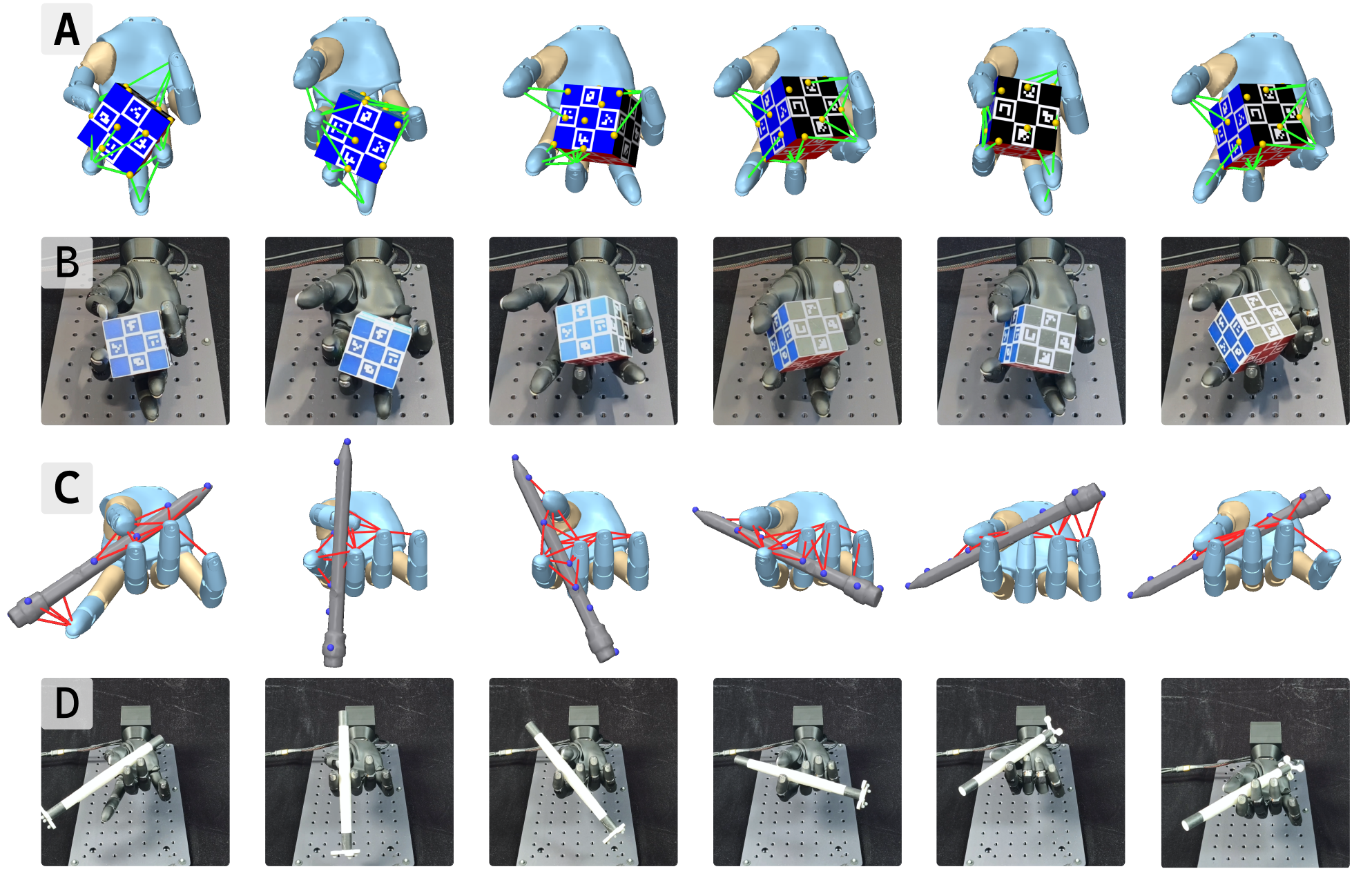}
    \vspace{-3mm}
    \caption{Interaction-mesh visualizations of retargeted reference trajectories (A,C) and corresponding zero-shot real-world executions (B,D) for cube reorientation and pen spinning on the Wuji Hand.}

    \label{fig:real_world_rollouts}
\end{figure}


As shown in Table~\ref{tab:retarget_tracking_combined}, \topomethod{} achieves the highest success rate at 84.4\% and the lowest position error at 0.87\,cm on the Ho-cap Dataset, while rotation errors remain comparable across methods because the trajectories are primarily grasp-centric. 
On the MoCap Pen-Spin Dataset, rapid motion and frequent non-tip contact transitions demand high-quality reference motion; \topomethod{} reaches 87.5\% success, over 40 percentage points above all baselines, with the lowest position and rotation errors. 
The widened performance gap on pen spinning highlights the applicability of \topomethod{} to contact-rich manipulation, where the active hand-object contact changes frequently over time.
Fig.~\ref{fig:real_world_rollouts} further shows zero-shot Wuji Hand sim-to-real transfer on cube reorientation and pen spinning, demonstrating that learned tracking policies can execute contact-rich skills with repeated in-hand contact changes.

\vspace{-2mm}
\subsection{Generalization and Augmentation without Per-Case Retuning}
\label{sec:generalization_aug}

\begin{figure}[htbp]
    \centering
    \includegraphics[width=0.8\textwidth]{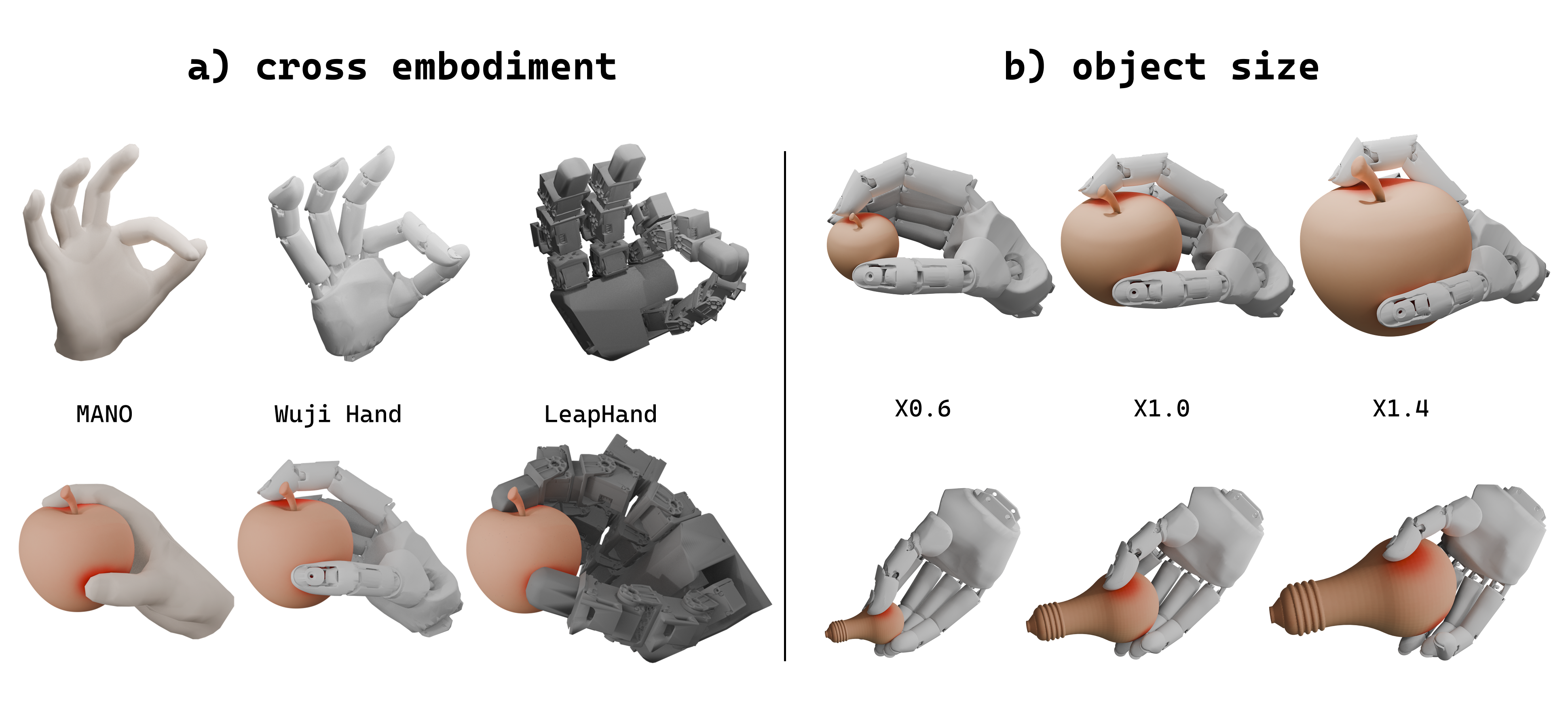}
    \vspace{-4mm}
    \caption{Augmentation across object scales and dexterous hand embodiments without per-case retuning.}

    \label{fig:generalization_aug}
\end{figure}

To answer Q3 on generalization and augmentation without per-case retuning, we retarget the same human demonstrations to different dexterous hand embodiments and object scales while keeping all retargeting parameters fixed.
Fig.~\ref{fig:generalization_aug} shows that the proposed approach transfers both the hand motion and the hand-object interaction under the embodiment and scale changes. 
The \teacheredit{interaction-mesh} formulation supports this transfer by rebuilding the local hand-object graph with edge weights recomputed from the scaled object mesh, providing a simple way to augment human demonstrations.



\section{Conclusion}
\label{sec:conclusion}

We presented \topomethod{}, a fixed-parameter retargeting framework that converts human hand-object demonstrations into dexterous robot references. \topomethod{} \teacheredit{maintains hand-object interaction in the object frame} through distance-weighted Laplacian optimization, directional consistency, kinematic constraints, and penetration handling. This design improves reference-level retargeting quality, achieving better contact precision, contact alignment, and penetration metrics than representative baselines on the ContactPose Dataset. The improved references also lead to stronger downstream RL tracking, with higher success rates and lower object tracking errors, especially on agile pen-spinning demonstrations with frequent non-tip contact transitions. The same formulation supports augmentation across object scales and hand embodiments without task-specific tuning.

\section{Limitations and Future Work}
\topomethod{} depends on the quality of the upstream human reference motion. Our method can handle contact distortion caused by penetration in the source motion, but it is less effective for virtual contacts, where the source finger is intended to interact with the object but does not actually touch the object surface. Future work can incorporate source-motion preprocessing to correct such missing-contact cases before retargeting.


\acknowledgments{%
We would like to thank Fan Yang, Chaoyi Pan, Cunxi Dai, and Xiaofeng Guo for valuable discussions and helpful feedback on the paper. We are also grateful to Tongbin Liu and Wenkai Nie for their support in preparing the experimental setup.
}


\clearpage


\bibliography{example}  

\clearpage

\appendix

\section{\teacheredit{Appendix}}


\subsection{Retargeting Details}
\label{app:retarget_details}

\noindent\textbf{Implementation detail for $E_{\mathrm{reg}}$}.
In the implementation, the regularization term in the final refinement is expanded as
\begin{equation}
\begin{aligned}
E_{\mathrm{reg}}(q;q_{t-1}^{r,*})
=&
\lambda_{\mathrm{reg}}\left\|q-q_{t-1}^{r,*}\right\|_2^2
+\lambda_{\mathrm{base,pos}}\left\|q_{\mathrm{pos}}^{\mathrm{base}}\right\|_2^2
+\lambda_{\mathrm{base,rot}}\left\|q_{\mathrm{rot}}^{\mathrm{base}}\right\|_2^2 .
\end{aligned}
\end{equation}
The first term enforces temporal smoothness in the optimizer coordinates, while the last two terms are a lightweight base-pose prior for floating-base retargeting.
Here, $q_{\mathrm{pos}}^{\mathrm{base}}$ and $q_{\mathrm{rot}}^{\mathrm{base}}$ denote the translational and rotational components of the robot base pose $q^{\mathrm{base}}$, respectively.

\begin{table}[htbp]
\centering
\caption{Selected retargeting parameter settings.}
\label{tab:supp_retarget_params}
\footnotesize
\setlength{\tabcolsep}{5pt}
\renewcommand{\arraystretch}{1.05}
\begin{tabularx}{\linewidth}{@{}llX@{}}
\toprule
\textbf{Parameter} & \textbf{Value} & \textbf{Description} \\
\midrule
$N_o$ & $50$ & Number of object surface samples in the interaction mesh. \\
$\lambda_{\mathrm{IM}}$ & $500$ & Weight of the interaction-mesh Laplacian objective. \\
$\kappa$ & $30$ & Distance-decay factor for the exponential Laplacian edge weights. \\
$\lambda_{\mathrm{warm}}$ & $1$ & Bone-direction weight in the initialization stage. \\
$\lambda_{\mathrm{bone}}$ & $0.1$ & Bone-direction prior weight in the final refinement stage. \\
$\lambda_{\mathrm{smooth}}$ & $2.5$ & Temporal smoothness weight in the initialization stage. \\
$\lambda_{\mathrm{reg}}$ & $2.5$ & Temporal smoothness weight inside $E_{\mathrm{reg}}$. \\
$\lambda_{\mathrm{base,pos}}$ & $100$ & Floating-base translation prior weight used in $E_{\mathrm{reg}}$. \\
$\lambda_{\mathrm{base,rot}}$ & $1$ & Floating-base rotation prior weight used in $E_{\mathrm{reg}}$. \\
$\tau$ & $0.001\,\mathrm{m}$ & Soft penetration tolerance. \\
$b$ & $0.030\,\mathrm{m}$ & Hard penetration backstop. \\
$w_s$ & $1.0\times 10^5$ & Slack penalty weight for the penetration constraint. \\
\bottomrule
\end{tabularx}
\end{table}

\subsection{OmniRetarget Baseline}
\label{app:omniretarget_baseline}
{\teachereditcolor
OmniRetarget~\cite{yang2025omniretarget} is originally designed for
humanoid whole-body loco-manipulation and scene interaction. For our
dexterous hand-object baseline, we keep its interaction-mesh objective
and object surface sampling, but replace the humanoid source keypoints
with MediaPipe hand keypoints:
\[
V_t^s = [P_t^{\mathrm{mp}}; O_t^s],
\qquad
V_t^r(q) = [P_t^r(q); O_t^r],
\]
where $P_t^{\mathrm{mp}}\in\mathbb{R}^{21\times 3}$ denotes the source
MediaPipe hand keypoints and $O_t^s,O_t^r$ are the sampled object points
under the source and retargeted object poses.

We make two hand-specific changes. First, the wrist is not fixed as a
root anchor; it is included in $P_t^r(q)$ and optimized jointly with the
other hand joints. Second, collision handling is applied to the full hand
geometry rather than to a sparse humanoid contact set, since dexterous
manipulation may contact the object with fingertips, intermediate
phalanges, finger sides, or the palm.
}

\subsection{Retargeting Metrics}
\label{app:retarget_metrics}
Let $\mathcal{C}$ denote the in-contact hand links, where each link $c$ associates a source hand point $h^s_c$, a retargeted robot point $h^r_c$, the corresponding source and retargeted object points $o^s_c$ and $o^r_c$, and a parent joint with source and retarget positions $h^s_{\mathrm{pa}(c)}$ and $h^r_{\mathrm{pa}(c)}$ (the wrist for the proximal link). The set $\mathcal{C}$ is obtained directly from the ContactPose contact annotations using ContactPose's own attribution: per-vertex contact intensity is sigmoid-normalized and thresholded, each contact vertex is assigned to its nearest of the $20$ hand bones by point-to-segment distance, and a link is marked in contact when at least $10$ vertices fall on it. Contact precision measures the object-relative position error:
\begin{equation}
\begin{aligned}
E_{\mathrm{prec}}
&=
\frac{1}{|\mathcal{C}|}
\sum_{c\in\mathcal{C}}
\left\|
\left(h^r_c-o^r_c\right)
-
\left(h^s_c-o^s_c\right)
\right\|_2 .
\end{aligned}
\label{eq:contact_precision}
\end{equation}
Contact alignment measures the angular deviation between the source and retargeted bone-segment orientations of the in-contact links:
\begin{equation}
\begin{aligned}
E_{\mathrm{align}}
&=
\frac{1}{|\mathcal{C}|}
\sum_{c\in\mathcal{C}}
\cos^{-1}
\left(
\operatorname{clip}
\left(
\hat u^r_c\cdot \hat u^s_c, -1, 1
\right)
\right),
\\
\hat u^r_c
&=
\frac{h^r_c-h^r_{\mathrm{pa}(c)}}{\|h^r_c-h^r_{\mathrm{pa}(c)}\|_2},
\qquad
\hat u^s_c
=
\frac{h^s_c-h^s_{\mathrm{pa}(c)}}{\|h^s_c-h^s_{\mathrm{pa}(c)}\|_2}.
\end{aligned}
\label{eq:contact_alignment}
\end{equation}
For penetration, let $\mathcal{X}^r_t$ be sampled robot-hand surface points and let $\mathcal{M}_t$ denote the posed object mesh with signed distance $d_{\mathcal{M}_t}(x)$, positive outside the object. We report the maximum penetration depth and the fraction of frames whose maximum penetration exceeds $\delta=2\,\mathrm{mm}$:
\begin{equation}
\begin{aligned}
D_{\mathrm{pen}}^{\max}
&=
\max_t\max_{x\in\mathcal{X}^r_t}
\left[-d_{\mathcal{M}_t}(x)\right]_+,
\\
R_{\mathrm{pen}}(\delta)
&=
\frac{1}{N}
\sum_{t=1}^{N}
\mathbf{1}
\left[
\max_{x\in\mathcal{X}^r_t}
\left[-d_{\mathcal{M}_t}(x)\right]_+
>
\delta
\right].
\end{aligned}
\label{eq:penetration_metrics}
\end{equation}

\subsection{Experimental Setup}
\label{sec:experimental_setup}
\textbf{Datasets and Tasks.}
For reference motion retargeting evaluation, we use the ContactPose Dataset~\cite{brahmbhatt2020contactpose} to measure hand-object contact precision, contact alignment, and penetration. We evaluate on 25 of the 28 ContactPose grasps, excluding three with deeply concave object geometry (mug, scissors, Utah teapot) that no method reliably handles. For downstream tracking evaluation, we use two trajectory sets. Ho-cap~\cite{wang2026ho} provides 32 grasp-centric hand-object references that test whether retargeted motions remain trackable under relatively stable contact conditions. The 32 trajectories are randomly drawn from Ho-cap's single-hand interactions to fit our training budget. Our self-collected MoCap Pen-Spin Dataset contains 32 pen-spinning demonstrations recorded with motion capture and the Wuji Glove, with trajectories averaging 12.4\,s and containing frequent non-tip contacts and contact-mode transitions. For both datasets, all retargeting methods are evaluated on the same human references and object geometry, and each generated reference set is used to train a separate policy under the identical RL setup. We further deploy the trained policies on the Wuji Hand for cube reorientation and pen spinning.

\textbf{Baselines and Metrics.}
We compare \topomethod{} with OmniRetarget~\cite{yang2025omniretarget}, Mink~\cite{Zakka_Mink_Python_inverse_2026}, DexPilot~\cite{handa2020dexpilot}, and GeoRT~\cite{yin2025geometric} under the same human references and object geometry when applicable. Reference-level metrics include contact precision, contact alignment, maximum penetration, penetration rate, and solve time; full definitions are provided in Appendix~\ref{app:retarget_metrics}. Policy metrics include success rate and object pose tracking error.


\subsection{RL Training}
\label{sec:single_hand_tracking_mdp}
\label{app:rl_details}

We formulate single-hand object manipulation as a finite-horizon MDP. The goal is to track a reference manipulation clip with a right dexterous hand while keeping the manipulated object aligned with the demonstrated motion. A reference clip is a sequence
\[
  \xi^{\mathrm{ref}}
  =
  \{q^{\theta,\mathrm{ref}}_k,\; q^{o,\mathrm{ref}}_k,\;
    p^{\mathrm{ref}}_{1:L,k}\}_{k=0}^{N-1},
\]
where $q^{\theta,\mathrm{ref}}_k$ denotes the reference finger joint positions, $q^{o,\mathrm{ref}}_k \in \mathrm{SE}(3)$ denotes the object pose expressed in the base frame, and $p^{\mathrm{ref}}_{1:L,k}$ denotes the reference positions of the tracked hand links. We use the base frame for reference quantities so that the policy learns the relative hand-object interaction instead of a global trajectory.

\subsubsection{State and action}

The simulator state contains the hand joint state, actuator state, contact state, and object pose and velocity. The policy does not observe the full state. At each control step $t$, the active reference index is $k_t$. The action $a_t$ is a residual joint-position command for the $n_\theta$ right-hand finger joints. The low-level target is centered at the current reference pose,
\[
  q^{\theta,\mathrm{tar}}_t = q^{\theta,\mathrm{ref}}_{k_t} + a_t.
\]
This residual parameterization lets the policy correct tracking errors while using the demonstration as a strong prior.

\subsubsection{Observation}

The policy observation has three parts. First, proprioception includes the current finger joint positions, joint velocities, and the previous action. Second, the object observation represents the current object pose in the base frame with a small set of axis points attached to the object. This avoids direct dependence on a particular quaternion parameterization while still encoding position and orientation. Third, the reference observation contains the current reference joint pose, object axis points, and tracked hand-link positions, plus short lookahead reference features from future frames. In our implementation we use lookahead offsets of 1, 3, and 5 control steps. During training we add small noise to proprioception and the observed object pose to improve robustness.

\subsubsection{Reference initialization and sampling}

We use reference initialization at every reset. A start frame $k_0$ is sampled uniformly from the reference trajectory, and the simulator is reset to the corresponding reference hand pose and object pose. The reference index then advances with the simulator control steps, $k_{t+1}=k_t+1$. Uniform sampling over the trajectory exposes the policy to all phases of the manipulation and avoids overfitting to the beginning of the clip. We optionally apply small reset noise to the object pose around the sampled reference state.

\subsubsection{Reward}

The reward is dominated by tracking terms for the object, hand links, and joints. For an error $e$ and scale $\sigma$, we use a Gaussian tracking kernel
\[
  \psi(e;\sigma) = \exp\left(-\|e/\sigma\|^2\right).
\]
The object term tracks a set of axis points attached to the object, which jointly encode its base-frame position and orientation,
\[
  r_{\mathrm{obj}}
  =
  \psi\left(
    \frac{1}{6}\sum_{m=1}^{6}
    \|u_{m,t} - u^{\mathrm{ref}}_{m,k_t}\|_2;
    \sigma_{\mathrm{obj}}
  \right),
\]
where $u_{m,t}$ and $u^{\mathrm{ref}}_{m,k_t}$ are current and reference object axis points in the base frame. The hand-link term tracks the positions of the palm and finger links,
\[
  r_{\mathrm{link}}
  =
  \frac{1}{L}\sum_{\ell=1}^{L}
  \psi(\|p_{\ell,t} -p^{\mathrm{ref}}_{\ell,k_t}\|_2;\sigma_{\mathrm{link}}),
\]
and the joint term tracks the normalized finger joint error,
\[
  r_{\mathrm{joint}}
  =
  \frac{1}{n_\theta}\sum_{j=1}^{n_\theta}
  \psi\left(
    \frac{q^{\theta}_{j,t} - q^{\theta,\mathrm{ref}}_{j,k_t}}
         {q^{\max}_j - q^{\min}_j};
    \sigma_{\mathrm{joint}}
  \right).
\]
The total reward is a weighted sum:
\[
  r_t =
    w_{\mathrm{obj}} r_{\mathrm{obj}}
  + w_{\mathrm{link}} r_{\mathrm{link}}
  + w_{\mathrm{joint}} r_{\mathrm{joint}}
  + w_{\mathrm{smooth}} r_{\mathrm{smooth}},
\]
where $r_{\mathrm{smooth}}=\|a_t-a_{t-1}\|_2^2+\|a_t-2a_{t-1}+a_{t-2}\|_2^2$. In practice, the largest weight is assigned to object tracking, followed by link, joint, and smoothness terms.

The reward weights and scales, together with the termination conditions, are summarized in Table~\ref{tab:supp_rl_mdp}.

\begin{table}[htbp]
\centering
\caption{Reference-tracking MDP. We define $\psi(e;\sigma)=\exp(-\|e/\sigma\|^2)$.}
\label{tab:supp_rl_mdp}
\footnotesize
\setlength{\tabcolsep}{4pt}
\renewcommand{\arraystretch}{1.05}
\begin{tabular}{@{}lll@{}}
\toprule
\textbf{Term} & \textbf{Expression / specification} & \textbf{Weight} \\
\midrule
Object & $\psi(\frac{1}{6}\sum_{m=1}^{6}\|u_m-u_m^{\mathrm{ref}}\|_2;\,0.04)$ & 8.0 \\
Link position & $\frac{1}{L}\sum_{\ell=1}^{L} \psi(\|p_{\ell}-p_{\ell}^{\mathrm{ref}}\|_2;\,0.025)$ & 1.0 \\
Joint position & $\frac{1}{n_\theta}\sum_{j=1}^{n_\theta} \psi(|q_j^\theta-q_j^{\theta,\mathrm{ref}}|/(q_j^{\max}-q_j^{\min});\,0.1)$ & 1.0 \\
Action smoothness & $\|a_t-a_{t-1}\|_2^2+\|a_t-2a_{t-1}+a_{t-2}\|_2^2$ & -0.01 \\
\midrule
Timeout & Episode reaches 20\,s & -- \\
Object unstable & Height $<0.06$\,m, linear velocity $>10$\,m/s, or angular velocity $>500$\,rad/s & -- \\
Object position error & $\|p_o-p_o^{\mathrm{ref}}\|_2>0.05$\,m & -- \\
Object orientation error & Orientation error $>45^\circ$ & -- \\
Object axis-point error & Any of the 6 object axis-point errors $>0.05$\,m & -- \\
\bottomrule
\end{tabular}
\end{table}

\subsubsection{Domain randomization}

To reduce overfitting to a single simulator instance, training randomizes object mass and center of mass, contact friction, actuator gains, joint damping and armature, encoder bias, and hand link inertial properties. We also apply intermittent external force and torque disturbances to the object. These perturbations are sampled at reset or during rollout and are not included in the policy observation.

The full randomization ranges are listed in Table~\ref{tab:supp_rl_dr}.

\begin{table}[htbp]
\centering
\caption{Domain randomization ranges used during tracking-policy training.}
\label{tab:supp_rl_dr}
\footnotesize
\setlength{\tabcolsep}{5pt}
\renewcommand{\arraystretch}{1.05}
\begin{tabular}{@{}lll@{}}
\toprule
\textbf{Randomized quantity} & \textbf{Range / setting} & \textbf{Mode} \\
\midrule
Joint-position observation noise & $\mathcal{N}(0, 0.02)$ rad & Step obs. \\
Joint-velocity observation noise & $\mathcal{N}(0, 0.05)$ rad/s & Step obs. \\
Object axis-point position noise & $\mathcal{N}(0, 0.002)$ m & Step obs. \\
Object axis-point orientation noise & $\mathcal{N}(0, 0.01)$ rad & Step obs. \\
Observation delay & 0--2 control steps & Step obs. \\
Reference reset: finger joints & $U[-0.02, 0.02]$ rad & Reset \\
Reference reset: object position & $U[-0.005, 0.005]$ m & Reset \\
Reference reset: object orientation & Axis-angle perturbation, angle $U[-0.03, 0.03]$ rad & Reset \\
Object COM offset & $[-0.003, 0.003]$ m & Startup \\
Robot friction scale & $[0.7, 1.3]$ & Startup \\
Robot collision-geometry scale & $[0.9, 1.1]$ & Startup \\
Object mass / inertia scale & $[0.4, 1.6]$ & Startup \\
PD stiffness scale & $[0.75, 1.5]$, log-uniform & Startup \\
PD damping scale & $[0.5, 2.0]$, log-uniform & Startup \\
Joint damping scale & $[0.3, 3.0]$, log-uniform & Startup \\
Joint armature scale & $[0.75, 1.3]$ & Startup \\
Joint friction-loss scale & $[0.5, 2.0]$ & Startup \\
Encoder bias & $[-0.01, 0.01]$ rad & Startup \\
Robot link inertia scale & $[0.4, 1.5]$ & Startup \\
Robot link mass scale & $[0.4, 1.5]$ & Startup \\
External object force disturbance & $U[-0.25, 0.25]$\,N, single-step impulse every $0.6$--$1.8$\,s & Rollout \\
External object torque disturbance & $U[-0.00375, 0.00375]$\,N\,m & Rollout \\
\bottomrule
\end{tabular}
\end{table}


\subsubsection{Policy network and optimization}

The policy and value networks and the PPO optimizer use the settings in Table~\ref{tab:supp_rl_ppo}.

\begin{table}[htbp]
\centering
\caption{PPO training and network hyperparameters.}
\label{tab:supp_rl_ppo}
\footnotesize
\setlength{\tabcolsep}{5pt}
\renewcommand{\arraystretch}{1.05}
\begin{tabular}{@{}ll@{}}
\toprule
\textbf{Parameter} & \textbf{Value} \\
\midrule
Parallel environments & 4096 \\
Simulation step & 0.01\,s \\
Decimation & 5 \\
RL control frequency & 20\,Hz \\
Reference trajectory frequency & 20\,Hz \\
Episode length & 20\,s (400 control steps) \\
Rollout length & 40 control steps / environment \\
Samples per PPO iteration & 163{,}840 \\
Actor network & MLP [512, 256, 128], ELU \\
Critic network & MLP [512, 512, 256, 128], ELU \\
Observation normalization & Enabled \\
Action distribution & Softplus Gaussian \\
Optimizer & Adam \\
Learning rate & $1\times10^{-4}$ \\
PPO epochs / minibatches & 4 / 32 \\
Entropy coefficient & 0.001 \\
Discount $\gamma$ & 0.99 \\
GAE $\lambda$ & 0.95 \\
\bottomrule
\end{tabular}
\end{table}

\end{document}